\documentclass{article}

\usepackage{fancyvrb}
\usepackage{graphicx}
\usepackage{hyperref}
\usepackage{gensymb}

\begin{document}

\title{Medical diagnosis as pattern recognition in a framework of information compression by multiple alignment, unification and search}

\author{J Gerard Wolff\footnote{Dr Gerry Wolff, BA (Cantab), PhD (Wales), CEng, MBCS (CITP); CognitionResearch.org, Menai Bridge, UK; \href{mailto:jgw@cognitionresearch.org}{jgw@cognitionresearch.org}; +44 (0) 1248 712962; +44 (0) 7746 290775; {\em Skype}: gerry.wolff; {\em Web}: \href{http://www.cognitionresearch.org}{www.cognitionresearch.org}.}}

\maketitle

\begin{abstract}

This paper describes a novel approach to medical diagnosis based on the SP theory of computing and cognition. The main attractions of this approach are: a format for representing diseases that is simple and intuitive; an ability to cope with errors and uncertainties in diagnostic information; the simplicity of storing statistical information as frequencies of occurrence of diseases; a method for evaluating alternative diagnostic hypotheses that yields true probabilities; and a framework that should facilitate unsupervised learning of medical knowledge and the integration of medical diagnosis with other AI applications.

\end{abstract}

{\em Keywords:} medical diagnosis, information compression, multiple alignment, SP theory, pattern recognition, causal reasoning.

\section{Introduction}

The problem of providing computational support for medical diagnosis has been approached from many directions including logical reasoning, fuzzy logic, set theory, rough set theory, if-then rules, Bayesian networks, classical parametric and non-parametric statistics, artificial neural networks, case-based reasoning, support vector machines, perceptrons, possibility theory, and more, as well as various aggregations or combinations of methods \cite{mangiameli_etal_2004}.

This paper describes a novel approach to diagnosis based on the SP theory of computing and cognition (described below). The main attractions of this approach are:

\begin{itemize}

\item A format for representing diseases that is simple and intuitive.

\item An ability to cope with errors and uncertainties in diagnostic information.

\item The simplicity of storing statistical information as frequencies rather than conditional probabilities.

\item A method for evaluating alternative diagnostic hypotheses that yields true probabilities.

\item A framework that should facilitate unsupervised learning of medical knowledge and the integration of medical diagnosis with other AI applications.

\end{itemize}

{\em It must be stressed that the primary purpose of this paper is conceptual}: to describe an approach to medical diagnosis that is significantly different from the main alternatives and with potential advantages compared with those alternatives. Although a prototype of the proposed new system exists, it is not yet a shrink-wrapped software system that is ready for immediate application. Like any other diagnostic system, the SP system must be equipped with a body of relevant knowledge and the creation of such a body of knowledge (by automatic learning or by `knowledge elicitation' from experts) is a major undertaking in its own right.

Key elements of the SP theory are first described, just sufficient for present purposes. Section \ref{application_section} describes how the theory may be applied to medical diagnosis, viewed as a process of pattern recognition. This section also discusses how the SP system relates to several aspects of the diagnostic process, including causal reasoning and the process of acquiring the knowledge that is needed for accurate diagnosis. Section \ref{alternatives_section} compares this new approach to medical diagnosis with some of the alternatives. The paper concludes with an outline of what still needs to be done in this programme of research and with a review of the main points that have been made.

\section{The SP theory}\label{sp_theory}

The SP theory grew out of a long tradition in psychology that many aspects of brain function may be understood as information compression (see, for example, \cite{attneave_1954,barlow_1969,wolff_1988,chater_1996}). It is based on principles of {\em minimum length encoding}\footnote{An umbrella term for `minimum message length encoding' and `minimum description length encoding'.} pioneered by Solomonoff \cite{solomonoff_1964}, Wallace and Boulton \cite{wallace_boulton_1968}, Rissanen \cite{rissanen_1978} and others (see also \cite{li_vitanyi_1997}). An overview of the theory is presented in \cite{wolff_sp_overview} and more detail may be found in other papers cited there (see also \cite{wolff_2001_igpl}).

The SP theory has been developed as an abstract model of {\em any} system for processing information, either natural or artificial. In broad terms, the system receives `New' information from its environment and transfers it to a repository of `Old' information.
At the same time, it tries to compress the information as much as
possible by finding patterns that match each other and merging or
`unifying' patterns that are the same.\footnote{The term `unification' in the SP theory means a simple merging of two or more identical patterns to make one. This meaning is different from but related to the meaning of the term in logic.} An important part of this process is the building of `multiple alignments' as described below.

The SP framework is Turing-equivalent in the sense that it can model a universal Turing machine \cite{wolff_1999_comp} but it has much more to say about the nature of `intelligence' than the Turing model of computing (or equivalent models such as lamda calculus \cite{church_1941} or the Post canonical system \cite{post_1943}).

To date, the main areas in which the SP framework has been applied are probabilistic reasoning, pattern recognition and information retrieval \cite{wolff_1999_prob}, parsing and production of natural language \cite{wolff_2000}, modelling concepts in logic and mathematics \cite{wolff_maths_logic}, and unsupervised learning \cite{wolff_cavtat_2003,wolff_unsupervised_learning}.

\subsection{Computer models}

Two computer models of the SP system have been developed:

\begin{itemize}

\item SP62 is a partial realisation of the theory that does not transfer any information from New to Old. This model tries to compress the New information in terms of the Old information by building multiple alignments of the kind that will be seen below. SP62 also contains procedures for calculating the probabilities of inferences that may be drawn from alignments. A slightly earlier version of this model (SP61) is described quite fully in \cite{wolff_2000}. Both versions are relatively robust and mature.

\item SP70 realises all the main elements of the theory, including the transfer of information from New to Old. In addition to building multiple alignments like SP62, the model compiles one or more alternative `grammars' for the information in New, using principles of minimum length encoding. This model, and its application to unsupervised learning, is described quite fully in \cite{wolff_cavtat_2003,wolff_unsupervised_learning}. More work is required to realise the full potential of this model.

\end{itemize}

\subsection{Representation of knowledge}\label{representation_of_knowledge}

In the SP system, {\em all} kinds of knowledge are stored as arrays of atomic {\em
symbols} in one or two dimensions called {\em patterns}. In work to date,
the main focus has been on one-dimensional patterns (i.e., sequences of symbols) but it is envisaged that, at some stage, the concepts will be generalised to patterns in two dimensions.

For present purposes, we may define patterns and symbols as follows:

\begin{itemize}

\item A {\em pattern} is a sequence of symbols bounded by end-of-pattern characters such as
`(' and `)', not shown in the examples in this paper.

\item A {\em symbol} is a string of non-space characters bounded by white space (space characters, line-feed characters and the like).

\item Any symbol can be matched with any other symbol and, for any one pair of symbols, the two symbols are either `the same' or `different'. No other result is permitted.

\item Symbols have no intrinsic meaning such as `add' for the symbol `+' in arithmetic or `multiply' for the symbol `$\times$'. Any meaning attaching to an SP symbol takes the form of one or more other symbols with which it is associated in a given set of patterns.

\item Each pattern has an associated integer value representing the absolute or relative frequency of occurrence of that pattern in some domain.

\end{itemize}

Despite the extraordinary simplicity of this format for representing knowledge, the way in which SP patterns are processed within the system means that they can model a wide variety of established representational schemes, including context-free and context-sensitive grammars, class-inclusion hierarchies, part-whole hierarchies, discrimination networks and trees, if-then rules, and others.

\subsection{Processing knowledge}\label{processing_knowledge}

A key part of the process of matching patterns is the building of `multiple alignments', described and illustrated here. The process of building multiple alignments in the SP system provides a unified model for a variety of computational effects including fuzzy pattern recognition, best-match information retrieval, probabilistic and exact styles of reasoning, unsupervised learning, planning, problem solving and others, as described in \cite{wolff_sp_overview}.

\subsubsection{Multiple alignments}

This subsection and the ones that follow describe the main elements of the multiple alignment concept as it has been developed in the SP theory and explains how multiple alignments are created and evaluated in the SP system.

In bioinformatics, a multiple alignment is an arrangement of two or more DNA sequences or sequences of amino acid residues so that matching symbols are aligned. Fig. \ref{DNA_figure} shows a typical example. The general idea is that, by judicious `stretching' of sequences, as many symbols as possible are aligned with each other. A variety of measures of the `goodness' of alignments are used but they all tend to favour alignments where the number of aligned symbols is high and the gaps between them are relatively few and relatively small.

\begin{figure}[!hbt]
\fontsize{10.00pt}{12.00pt}
\centering
\begin{BVerbatim}
  G G A     G     C A G G G A G G A     T G     G   G G A
  | | |     |     | | | | | | | | |     | |     |   | | |
  G G | G   G C C C A G G G A G G A     | G G C G   G G A
  | | |     | | | | | | | | | | | |     | |     |   | | |
A | G A C T G C C C A G G G | G G | G C T G     G A | G A
  | | |           | | | | | | | | |   |   |     |   | | |
  G G A A         | A G G G A G G A   | A G     G   G G A
  | |   |         | | | | | | | |     |   |     |   | | |
  G G C A         C A G G G A G G     C   G     G   G G A
\end{BVerbatim}
\caption{A `good' alignment amongst five DNA sequences.}
\label{DNA_figure}
\end{figure}

In the SP framework, the concept of multiple alignment has been modified as follows:

\begin{itemize}

\item One or more of the sequences (termed {\em patterns}, as described in Section \ref{representation_of_knowledge}) are classified as `New' and the rest are `Old'.

\item A `good' alignment is one where the New patterns can be encoded economically in terms of the Old patterns in the alignment, as will be explained.

\item Any one pattern may appear more than once in one alignment, not as two or more {\em copies} but as two or more {\em appearances} of a {\em single} pattern. This has implications for the way alignments are formed and for the representation of recursive structures. These aspects of the multiple alignment concept are not relevant to the main proposals here and will not be considered further in this paper. Readers who wish to know more may consult the sources cited earlier.

\end{itemize}

Normally, the SP62 model is run with relatively few New patterns and a relatively large `database' or `dictionary' of Old patterns. The system typically forms a set of {\em alternative} alignments, each one of which represents a possible encoding of the New pattern or patterns in terms of one or more of the Old patterns.

Fig. \ref{spelling_alignment} shows a simple example, with one New pattern (in row 0, representing a badly-spelled version of the word `experimentation') and one Old pattern (in row 1, representing the correctly-spelled version of the word).\footnote{By convention, the New pattern or patterns are always shown in row 0 of alignments like those shown in Figs.
\ref{spelling_alignment} and \ref{parsing_alignment}, and the Old patterns are shown in the other rows, one pattern per row and in an order that is entirely arbitrary, without special significance. As we shall see, alignments can sometimes fit better on the page if they are rotated by $90\degree$ and in this case the New pattern or patterns are shown in column 0 with the Old patterns in the other columns, one pattern per column.} This is the best alignment produced by SP62 with the New pattern as shown and a dictionary of Old patterns, each one of which represents one word in its correctly-spelled form.

\begin{figure}[!hbt]
\fontsize{10.00pt}{12.00pt}
\centering
\begin{BVerbatim}
0      e   p e r i m a n t a t x p i u n   0
       |   | | | | |   | | | |     |   |
1 < E3 e x p e r i m e n t a t     i o n > 1
\end{BVerbatim}
\caption{The best alignment formed by SP62 when it is supplied with one New pattern (`e p e r i m a n t a t x p i u n') and a dictionary of Old patterns, each one of which represents a correctly-spelled word.}
\label{spelling_alignment}
\end{figure}

Fig. \ref{parsing_alignment} is a slightly more complicated example, the best alignment produced by SP62 when it is supplied with one New pattern (`j o h n r u n s') and a set of Old patterns, each one of which represents a grammatical rule. This alignment shows how the sentence `j o h n r u n s' (in row 0) may be analysed (`parsed') into its parts. The Old patterns in rows 1 to 3 represent grammatical rules: `$<$ S $<$ N $>$ $<$ V $>$ $>$' in row 3 means that a (simple) sentence is composed of a noun (`N') followed by a verb (`V'), `$<$ N 0 j o h n $>$' in row 2 means that `j o h n' is a noun, and `$<$ V 1 r u n s $>$' in row 1 means that `r u n s' is a verb.

\begin{figure}[!hbt]
\fontsize{10.00pt}{12.00pt}
\centering
\begin{BVerbatim}
0           j o h n         r u n s     0
            | | | |         | | | |
1           | | | |   < V 1 r u n s >   1
            | | | |   | |           |
2     < N 0 j o h n > | |           |   2
      | |           | | |           |
3 < S < N           > < V           > > 3
\end{BVerbatim}
\caption{The best alignment formed by SP62 when it is supplied with one New pattern (`j o h n r u n s') and a set of Old patterns, each one of which represents a grammatical rule.}
\label{parsing_alignment}
\end{figure}

\subsubsection{Evaluation of alignments}\label{evaluation_of_alignments}

As previously mentioned, a `good' multiple alignment is one where the New pattern or patterns can be encoded economically in terms of the Old patterns in the alignment. How is this evaluation done?

Every symbol has an associated `weight' which is the number of bits needed to encode that symbol. And the weight is derived from the frequency of occurrence of the symbol using the Shannon-Fano-Elias method (see \cite{cover_thomas_1991}) (a method that is similar to the well-known Huffman method). The frequency value for any symbol is derived from the frequency value of the pattern (or patterns) in which that symbol appears (as described in Section \ref{representation_of_knowledge}). In the context of medical diagnoses, the frequency associated with any given pattern is the frequency of occurrence of the disease that is represented by that pattern (see Section \ref{diseases_as_patterns}, below).

A few of the symbols (normally one or two) near the beginning of each Old pattern are classified as {\em identification} symbols or `ID-symbols'. For example, the ID-symbol in the pattern `$<$ E3 e x p e r i m e n t a t i o n $>$' in Fig. \ref{spelling_alignment} is `E3', and the ID-symbols in the pattern `$<$ N 0 j o h n $>$' in Fig. \ref{parsing_alignment} are `N' and `0'.

A `code' for any alignment may be derived quite simply by scanning the alignment from left to right looking for columns in the alignment that contain {\em one} code symbol, not matched with any other identical symbol. The code for the alignment is the sequence of symbols that have been found, in the given order. For example, the code derived in this way from the alignment in Fig. \ref{spelling_alignment} is `E3' and the code derived from the alignment in Fig.
\ref{parsing_alignment} is `S 0 1'.

The next step in the evaluation of a given alignment is the calculation of a `compression score' or `compression difference' as:

\begin{equation}
CD = B_n - B_e,
\label{equation_1}
\end{equation}

\noindent where $B_n$ is the total size (in bits) of those symbols within the New pattern that have been matched to Old symbols within the alignment, and $B_e$ is the total size (in bits) of the symbols in the code that has been derived as just explained. Adjustments to this score are made to take account of gaps in the alignment like those that may be seen in Fig. \ref{spelling_alignment}. The details of these calculations and adjustments are explained in \cite{wolff_2000}.

\subsubsection{The building of multiple alignments}

In bioinformatics, it is generally understood that the abstract `space' of alternative possible alignments between two or more sequences is, with few exceptions, astronomically large---which means that it cannot be searched exhaustively. All practical methods for finding `good' alignments amongst two or more sequences use heuristic methods such as `hill climbing (or `descent'), `beam search', `genetic algorithms' or the like that search selectively and exclude large parts of the search space. With methods like these, one can find good approximate solutions in a reasonable time but one can never be sure of finding the best possible solution (unless the sequences are very short and very few). Finding good multiple alignments in the SP system is no different.

At the heart of the SP system for building multiple alignments is an improved version of `dynamic programming' for finding full matches and good partial matches between two patterns (see, for example, \cite{sankoff_kruskall_1983}). Unlike standard versions of dynamic programming, the procedure used in the SP models:

\begin{itemize}

\item Can find good matches between patterns without restrictions on the lengths of the patterns.

\item Can normally find several {\em alternative} alignments between two patterns, not just one.

\item It allows the `depth' or thoroughness of searching to be varied according to need.

\end{itemize}

Given one New pattern and a database of Old patterns, SP62 first builds a set of alignments, each one of which is between the New pattern and {\em one} of the Old patterns. From this set, it selects the best few alignments, using the measure described in Section \ref{evaluation_of_alignments}. Each of these alignments can itself be treated as if it was a single pattern. So, in the next stage, SP62 builds larger alignments, each one of which is between one of the selected alignments and one of the Old patterns or between one of the selected alignments and another of those alignments. As before, the program selects the best of the alignments that have been formed.

The process is repeated in this way until no more alignments can be found. The process for building alignments containing two or more New patterns is a generalisation of what has been described here.

\subsubsection{Unsupervised learning}\label{unsupervised_learning}

At its most abstract level (Section \ref{sp_theory}), the SP model is conceived as a system that learns by transferring New information to its repository of Old information and compressing it at the same time. This abstract conception has now been realised more concretely in the form of the SP70 computer model \cite{wolff_cavtat_2003,wolff_unsupervised_learning} that is capable of learning simple grammars from raw data. However, further development of the model is needed to realise its full potential.

The current model has two stages:

\begin{enumerate}

\item From partial alignments between patterns, the model creates new patterns that are added to the repository of Old patterns as explained below.

\item Amongst the patterns that are generated in this way, some are `good' in terms the principles of minimum length encoding and others are `bad'. In the second stage of processing, the model measures the frequency with which each pattern may be recognised in the raw data and then it uses this information in a hill-climbing search amongst subsets of the Old patterns to find one or more sets of patterns that are good in terms of the principles of minimum length encoding. The remaining patterns may be discarded.

\end{enumerate}

It is envisaged that, when the model is more fully developed, these two stages will be repeated so that the system can progressively bootstrap a set of patterns that are good in terms of the principles of minimum length encoding, and thus represent a distillation of the patterns of redundancy in the original data.

SP70 is currently targeted at the learning of syntax in natural languages. Given a partial alignment like this:

\begin{center}
\begin{BVerbatim}
0      t h e g i r l r u n s   0
       | | |         | | | |
1 < %1 t h e b o y   r u n s > 1
\end{BVerbatim}
\end{center}

\noindent the program creates patterns like these:

\begin{center}
\begin{BVerbatim}
< %2 t h e >
< %3 r u n s >
< %4 0 b o y >
< %4 1 g i r l >
< %5 < %2 > < %4 > < %3 > >
\end{BVerbatim}
\end{center}

The first four are derived from coherent sequences of matched symbols and coherent sequences of unmatched symbols in the alignment and they correspond to what we would normally recognise as words. Each one has a grammatical category represented by ID-symbols such as `\%2', `\%3' and `\%4'. Notice that `b o y' and `g i r l' belong to the same disjunctive category \{`b o y', `g i r l'\} because they are alternatives at the same point in the original alignment and they both share the ID-symbol `\%4'. The ID-symbols `0' and `1' serve to distinguish the two alternatives in that category.

The last pattern in this example ties everything together by listing the sequence of categories in the original alignment. It is an `abstract' pattern describing the overall structure of the two original sentences.

For an application like medical diagnosis, the style of learning just described is probably not entirely appropriate. Section \ref{automatic_learning} describes how similar principles may be applied to medical data. If the potential of these ideas can be realised, the SP system should facilitate the automatic or semi-automatic construction of knowledge bases from raw medical data.

\subsubsection{Computational complexity}\label{computational_complexity}

The time complexity of the SP62 model in a serial processing environment is approximately O$(log_2 n \times nm)$, where $n$ is the size of the New pattern or patterns (in bits) and $m$ is the total size of the patterns in Old (in bits). In a parallel processing environment, the time complexity may approach O$(\log_2 n \times n)$, depending on how well the parallel processing is applied. The space complexity in serial or parallel environments is O$(m)$. Further details may be found in \cite{wolff_2000}.

In medical diagnosis, it seems reasonable to suppose that there will normally be a fairly small maximum for the number of signs and symptoms (abbreviated hereinafter as `symptoms') exhibited by any one patient. Correspondingly, there should be a maximum size for the size of the set of New patterns that are used to represent the patient's symptoms. If we take this to be a constant value for $n$, then in a serial processing environment the time complexity is approximately O$(m)$ and in a parallel processing environment it may approach O(1).

\section{Application of the SP system to medical diagnosis}\label{application_section}

To a large extent, medical diagnosis may be viewed as a problem of (fuzzy) pattern recognition: finding the best fit between a given set of symptoms for an individual patient and the symptoms associated with one or more diseases. However, causal reasoning also has a part to play when, for example, it is understood that a given disease is caused by a bacterial or virus infection.

This section presents an example showing how the SP system may be applied to medical diagnosis, viewed as a process of pattern recognition. The system may also support causal reasoning about medical problems and this is discussed briefly in Section \ref{causal_reasoning}, below.

Other aspects of the proposals are discussed in other subsections.

\subsection{Describing diseases using SP patterns}\label{diseases_as_patterns}

In the SP scheme, knowledge about diseases may be stored as patterns in a repository of Old information, and the symptoms for an individual patient may be represented as a set of one or more New patterns.

A pattern in the store of Old information may represent one disease and its associated symptoms, or a combination of diseases (see Section \ref{disease_combinations}, below), or it may represent a cluster of symptoms that tend to occur together in two or more different diseases (see Section \ref{dereferencing_of_pointers}, below). In addition, Old may include patterns that play supporting r{\^o}les (see Section \ref{uncertainties_in_diagnosis}, below).

The frequency value of each pattern may be used to represent the absolute or relative frequency with which a given disease or cluster of symptoms is found in a given population. These figures may be derived from epidemiological surveys or they may be estimated by medical experts.

By way of illustration, Fig. \ref{patterns_1} shows five examples of such patterns, one describing the symptoms of chicken pox, another describing the symptoms of smallpox, the third describing the cluster of symptoms which is described as `fever' and two more describing the class of `high' temperatures (38--39${^o}$C and 40+${^o}$C). The number in brackets after each pattern is a very rough estimate of the relative frequencies of occurrence of the corresponding disease or condition.\footnote{The figure for smallpox is clearly too high in the world today but it will serve for the purpose of illustration.}

\begin{figure}[!hbt]
\fontsize{09.00pt}{10.80pt}
\centering
\begin{BVerbatim}
<disease> chicken_pox :
     <dname> Chicken_Pox </dname>
     <R2> fever </R2>
     <appetite> normal </appetite>
     <chest> normal </chest>
     <chills> no </chills>
     <cough> no </cough>
     <diarrhoea> no </diarrhoea>
     <fatigue> no </fatigue>
     <lymph_nodes> normal </lymph_nodes>
     <malaise> yes </malaise>
     <muscles> normal </muscles>
     <nose> normal </nose>
     <skin> rash </skin>
     <throat> normal </throat>
     <weight_change> no </weight_change>
     <causative_agent> chicken_pox_virus </causative_agent>
     <treatment> chicken_pox_treatment </treatment>
</disease> (2500)

<disease> smpx :
     <dname> Smallpox </dname>
     <R1> flu_symptoms </R1>
     <appetite> normal </appetite>
     <chest> normal </chest>
     <diarrhoea> no </diarrhoea>
     <fatigue> no </fatigue>
     <lymph_nodes> normal </lymph_nodes>
     <malaise> no </malaise>
     <skin> rash with blisters </skin>
     <weight_change> no </weight_change>
     <causative_agent> smallpox_virus </causative_agent>
     <treatment> smallpox_treatment </treatment>
</disease> (5)

<disease> fever
     <breathing> rapid </breathing>
     <face> flushed </face>
     <temperature> <t1> </t1> </temperature>
</disease> (15000)

<t1> 38-39 </t1> (14705)
<t1> 40+ </t1> (147)
\end{BVerbatim}
\caption{\small Five SP patterns, one describing the symptoms of chicken pox, another describing the symptoms of smallpox, the third describing the symptoms of fever and two more describing the class of `high' temperatures.}
\label{patterns_1}
\end{figure}

\sloppy Each of the first three patterns begins and ends with a pair of symbols `$<$disease$>$ ... $<$/disease$>$' which indicate that the pattern describes a disease or a cluster of disease symptoms. Within each pattern, there are similar pairs of symbols, each one marking the beginning and end of a `field' which describes some aspect of the disease or cluster. For example, `$<$dname$>$ Chicken\_Pox $<$/dname$>$' provides the name of the chicken pox disease, `$<$skin$>$ rash $<$/skin$>$' describes one of its symptoms, `$<$causative\_agent$>$ chicken\_pox\_virus $<$/causative\_agent$>$' describes what causes the disease, and `$<$treatment$>$ chicken\_pox\_treatment $<$/treatment$>$' is a remarkably unhelpful description of how to treat the disease which would, of course, be much more detailed in a fully developed knowledge base.

Within the pattern for chicken pox, the field `$<$R2$>$ fever $<$/R2$>$' indicates that `fever' is one of the symptoms of the disease. However, by contrast with other fields like those just mentioned, the symbol `fever' is, in effect, a reference or pointer to a cluster of symptoms such as rapid breathing, flushed face and high temperature described in the third pattern in the same figure. In a similar way, `$<$R1$>$ flu\_symptoms $<$/R1$>$' in the pattern for smallpox is a reference or pointer to another pattern, not shown in the figure, that describes a cluster of symptoms associated with influenza and flu-like diseases.  The way in which pointers like these are dereferenced in the SP system will be seen in the next section.

Readers who are familiar with XML \cite{w3c_xml_spec} will notice that pairs of symbols like `$<$disease$>$ ... $<$/disease$>$' or `$<$dname$>$ ... $<$/dname$>$' are rather like the start and end tags used to mark the elements of an XML document. However, by contrast with XML and related languages such as HTML, symbols of that kind have no formal status in the SP system and the styles of symbols are not defined within the system. Any convenient style may be used such as `disease ... \#disease' or `disease ... \%disease' and in some applications it is not necessary to provide any distinctive markers for the beginnings and ends of patterns or fields. The concept of `field' has no formal status in the SP system either.

\subsection{Multiple alignment and medical diagnosis}\label{ma_medical_diagnosis}

The process of diagnosis may be modelled by the building of one or more multiple alignments. Fig. \ref{alignment_1} shows the best alignment created by SP62 with a set of New patterns shown in Fig. \ref{patterns_2} that describe `John Smith' and his symptoms and a set of Old patterns like those shown in Fig. \ref{patterns_1} that represent diseases or aspects of diseases.\footnote{Compared with the alignments shown in Figs. \ref{spelling_alignment} and \ref{parsing_alignment}, the alignment in Fig. \ref{alignment_1} has been rotated by $90{^o}$ to allow the alignment to fit better on the page. As previously noted, the New patterns are shown in column 0 and the Old patterns are shown in the other columns, one pattern per column in an order that is arbitrary and without special significance.}

\begin{figure}[!hbt]
\fontsize{09.00pt}{10.80pt}
\centering
\begin{BVerbatim}
<patient> John_Smith </patient>
<face> flushed </face>
<appetite> poor </appetite>
<breathing> rapid </breathing>
<muscles> aching </muscles>
<chills> yes </chills>
<fatigue> yes </fatigue>
<lymph_nodes> normal </lymph_nodes>
<malaise> no </malaise>
<nose> runny </nose>
<temperature> 38-39 </temperature>
<throat> sore </throat>
\end{BVerbatim}
\caption{\small The set of New patterns supplied to SP62 for the example discussed in the text. These patterns represent the patient `John Smith' and his symptoms.}
\label{patterns_2}
\end{figure}

\begin{figure}[!hbt]
\fontsize{06.00pt}{07.20pt}
\centering
\begin{BVerbatim}
0                1                    2                    3              4                5

                 <disease> ---------- <disease> ---------- <disease> ---- <disease>
                                      flu
                 : ------------------ :
<patient> ------ <patient>
John_Smith
</patient> ----- </patient>
                 <dname> ------------ <dname>
                                      Influenza
                 </dname> ----------- </dname>
                 <R1> --------------- <R1>
                                      flu_symptoms ------- flu_symptoms
                 </R1> -------------- </R1>
                 <R2> ------------------------------------ <R2>
                                                           fever -------- fever
                 </R2> ----------------------------------- </R2>
<appetite> ----- <appetite> --------- <appetite>
poor                                  normal
</appetite> ---- </appetite> -------- </appetite>
<breathing> ---- <breathing> -------------------------------------------- <breathing>
rapid ------------------------------------------------------------------- rapid
</breathing> --- </breathing> ------------------------------------------- </breathing>
                 <chest> ------------ <chest>
                                      normal
                 </chest> ----------- </chest>
<chills> ------- <chills> -------------------------------- <chills>
yes ------------------------------------------------------ yes
</chills> ------ </chills> ------------------------------- </chills>
                 <cough> --------------------------------- <cough>
                                                           yes
                 </cough> -------------------------------- </cough>
                 <diarrhoea> -------- <diarrhoea>
                                      no
                 </diarrhoea> ------- </diarrhoea>
<face> --------- <face> ------------------------------------------------- <face>
flushed ----------------------------------------------------------------- flushed
</face> -------- </face> ------------------------------------------------ </face>
<fatigue> ------ <fatigue> ---------- <fatigue>
yes                                   no
</fatigue> ----- </fatigue> --------- </fatigue>
                 <headache> ------------------------------ <headache>
                                                           yes
                 </headache> ----------------------------- </headache>
<lymph_nodes> -- <lymph_nodes> ------ <lymph_nodes>
normal ------------------------------ normal
</lymph_nodes> - </lymph_nodes> ----- </lymph_nodes>
<malaise> ------ <malaise> ---------- <malaise>
no ---------------------------------- no
</malaise> ----- </malaise> --------- </malaise>
<muscles> ------ <muscles> ------------------------------- <muscles>
aching --------------------------------------------------- aching
</muscles> ----- </muscles> ------------------------------ </muscles>
<nose> --------- <nose> ---------------------------------- <nose>
runny ---------------------------------------------------- runny
</nose> -------- </nose> --------------------------------- </nose>
                 <skin> ------------- <skin>
                                      normal
                 </skin> ------------ </skin>
<temperature> -- <temperature> ------------------------------------------ <temperature>
                                                                          <t1> ----------- <t1>
38-39 ------------------------------------------------------------------------------------ 38-39
                                                                          </t1> ---------- </t1>
</temperature> - </temperature> ----------------------------------------- </temperature>
<throat> ------- <throat> -------------------------------- <throat>
sore ----------------------------------------------------- sore
</throat> ------ </throat> ------------------------------- </throat>
                 <weight_change> ---- <weight_change>
                                      no
                 </weight_change> --- </weight_change>
                 <causative_agent> -- <causative_agent>
                                      flu_virus
                 </causative_agent> - </causative_agent>
                 <treatment> -------- <treatment>
                                      flu_treatment
                 </treatment> ------- </treatment>
                 </disease> --------- </disease> --------- </disease> --- </disease>

0                1                    2                    3              4                5
\end{BVerbatim}
\caption{\small The best alignment found by SP62 with the set of patterns from Fig. \ref{patterns_2} in New (describing the symptoms of the patient `John Smith') and a set of patterns in Old describing a range of different diseases and named clusters of symptoms, together with the `framework' pattern shown in column 1.}
\label{alignment_1}
\end{figure}

An alignment like this may be interpreted as the result of a process of recognition. In this case, the symptoms that have been recognised are those of influenza, as shown in column 2. The following subsections discuss aspects of the alignment and of this interpretation.

\subsection{A `framework' pattern}

\sloppy In an application like this, it is convenient but not essential to include amongst the Old patterns a `framework' pattern like the one shown in column 1. This is a generalised pattern for diseases of all kinds that lists the main categories associated with diseases such as `$<$dname$>$ $<$/dname$>$' (the name of the disease), `$<$breathing$>$ $<$/breathing$>$' (the state of the patient's breathing) and `$<$temperature$>$ $<$/temperature$>$' (the patient's temperature), but it does not specify specific values for any category.

This framework pattern serves as an anchor point for symbols in other patterns and facilitates the formation of multiple alignments in accordance with the rules described in \cite{wolff_sp_overview} and earlier publications.

\subsection{The ordering of descriptors}

In an application like medical diagnosis, it is not obvious that there is any intrinsic order to the symptoms of a disease or associated descriptors such as the name of the patient. In describing a patient's symptoms, it should make no difference whether `high temperature' is mentioned before `runny nose' or the other way round.

In the SP framework, each pattern that describes a disease or a cluster of symptoms necessarily imposes an order in which categories of descriptors are specified. However, users of the system may specify the patient's symptoms in any order that is convenient. This is because symptoms are described using a set of New patterns and there is no intrinsic order amongst the New patterns supplied to the system. In our example, New patterns were supplied to the SP62 model in the order shown in Fig. \ref{patterns_2} but in the alignment shown in Fig. \ref{alignment_1} they appear in a completely different order.

Notice that this freedom in the ordering of descriptors only applies to whole patterns. When two or more symbols in one pattern are matched to two or more symbols in another, the order of the symbols in one pattern must be the same as the order of the matching symbols in the other pattern.

\subsection{Dereferencing of pointers}\label{dereferencing_of_pointers}

As already noted, a symbol like `fever' or `flu\_symptoms' in one pattern may serve as a reference or pointer to another pattern that describes a cluster of symptoms that may be found in two or more different diseases.

In Fig. \ref{alignment_1}, we can see how such pointers are `de-referenced' in the SP system. The symbol `flu\_symptoms' in column 2 is matched to the same symbol in column 3 where flu-like symptoms are listed. Likewise, the symbol `fever' in column 3 is matched to the same symbol in column 4 where the symptoms of fever or listed. Fever is itself part of the cluster of flu-like symptoms.

The provision of named clusters like these saves the need to specify the corresponding symptoms redundantly in each of the diseases where such clusters appear.

\subsection{Uncertainties in diagnosis}\label{uncertainties_in_diagnosis}

Diagnosis is not an exact process:

\begin{itemize}

\item Most diseases are `family resemblance' or `polythetic' concepts because the majority of symptoms associated with any given disease are neither necessary nor sufficient for the diagnosis of the disease: they are `characteristic' of the disease in the sense that any one such symptom need not be present in every case and any of them may be associated with other diseases.

\item There may be and frequently are errors in the observation or recording of symptoms.

\end{itemize}

SP62 can accommodate these kinds of uncertainty in diagnosis in two distinct ways:

\begin{itemize}

\item Because it looks for a global best match amongst patterns, it does not depend on the presence or absence of any particular symptom. Notice how SP62 has succeeded in constructing the alignment shown in Fig. \ref{alignment_1} despite there being no match for `poor' in
the New pattern `$<$appetite$>$ poor $<$/appetite$>$' and `yes' in the New pattern `$<$fatigue$>$ yes $<$/fatigue$>$' and no match for many of the symbols in the Old patterns.

Although the system does not depend on the presence or absence of any one symptom, particular symptoms can have a major impact on diagnosis, as described in Section \ref{explaining_away}, below.

\item Within the SP framework, it is not necessary for every symptom of a disease to be recorded as a specific value. For example, in column 4 of the alignment in Fig. \ref{alignment_1}, the pair of symbols `$<$t1$>$ $<$/t1$>$' represents a set of alternative values for the temperature associated with fever. In this case, there are just two values, represented in Old by the patterns `$<$t1$>$ 38-39 $<$/t1$>$' (high temperature) and `$<$t1$>$ 40+ $<$/t1$>$' (very high temperature), as shown at the bottom of Fig. \ref{patterns_1}. The first of these patterns is shown in column 5 of the alignment, matched to the temperature of the patient shown in column 0.

Being able to specify symptoms as sets of alternative values allows the system to accommodate the kind of variability which is so prominent in many diseases.

\end{itemize}

\subsection{Weighing alternative hypotheses and the calculation of probabilities}\label{calculation_of_probabilities}

In medical diagnosis, it is quite usual for the physician to consider alternative hypotheses about what disease or diseases the patient may be suffering from. The SP framework provides a model for this process in the way the system builds alternative alignments for any given pattern or set of patterns in New. Alignments---and the corresponding diagnoses---may be evaluated as follows.

As previously noted (Section \ref{evaluation_of_alignments}), a `compression difference' is calculated for each alignment as shown in Equation \ref{equation_1}. The value $B_e$ that is used in that equation may be translated into an {\em absolute} probability for the given alignment:

\begin{equation}
P = 2^{-B_e}.
\label{equation_2}
\end{equation}

\noindent For any one alignment (the $j$th alignment) in a set of alternative alignments, $a_1 ... a_n$, that encode the same symbols from New, a {\em relative} probability may be calculated as:

\begin{equation}
p_j = P_j / \sum_{i = 1}^{i = n} P_i.
\label{equation_3}
\end{equation}

\noindent A fuller account of the way probabilities are calculated may be found in \cite{wolff_1999_prob}.

Given that the New patterns represent the symptoms of one patient at a particular time and given that each pattern in Old describes a single disease or a single cluster of symptoms that may form part of the description of one or more diseases, then each alignment formed by SP62 represents an hypothesis about any {\em one} disease that the patient may have.

Where alternative alignments encode different subsets of the symbols in New, it is possible that the patient may be suffering from two or more diseases at the same time. This possibility is discussed in Section \ref{disease_combinations}, below. However, where two or more of the best alignments encode exactly the same symbols from New, then they represent {\em alternative} diagnostic hypotheses and they may be compared using values for relative probability ($p$).

When SP62 formed the alignment shown in Fig. \ref{alignment_1}, it also formed a similar alignment, matching exactly the same symbols in New, in which column 2 contained a pattern representing the symptoms of smallpox, instead of the pattern for influenza. The relative probability values calculated in this case were 0.99950 for influenza and 0.00049 for smallpox, reflecting the prevalence of those two diseases in the world today.\footnote{There are, of course, other factors that may be relevant---such as the possibility that someone might release the smallpox virus deliberately---but in this example, knowledge of such other factors has been excluded.}

\subsubsection{`Explaining away'}\label{explaining_away}

The symptoms of influenza and smallpox are quite similar, except for the very distinctive rash and blisters that occur in smallpox. The example shown in Fig. \ref{alignment_1} is silent about whether John Smith had a rash and blisters or not. If a rash and blisters had been seen to be absent, this would have been represented as `$<$skin$>$ normal $<$/skin$>$'. Given this lack of information about the state of the patient's skin, he may have either influenza or smallpox but he is very much more likely to have the former than the latter, as indicated by the calculated probabilities.

If `$<$skin$>$ rash\_with\_blisters $<$/skin$>$' is added to the symptoms recorded in New, and if SP62 is run again with the augmented set of symptoms, the best alignment found by the system is similar to that shown in Fig. \ref{alignment_1} but with the pattern for smallpox (the second pattern in Fig. \ref{patterns_1}) instead of the pattern for influenza in column 2 and with a match shown between `$<$skin$>$ rash\_with\_blisters $<$/skin$>$' in the set of New patterns and the same symbols in the pattern that describes smallpox. However, in this case {\em there is no other alignment that matches the same symbols in New}. Consequently, the relative probability of the best alignment is 1.0. In short, the addition of one distinctive symptom to the list of symptoms has a dramatic effect on the relative probabilities calculated by the system. Instead of a vanishingly small probability for smallpox (0.00049), the system now assigns it a probability of 1.0, in accordance with our intuitions.

From this result, we may conclude that the patient certainly has smallpox and that his aching muscles and runny nose are due to smallpox, not influenza. This is the phenomenon of `explaining away': ``If A implies B, C implies B, and B is true, then finding that C is true makes A {\em less} credible. In other words, finding a second explanation for an item of data makes the first explanation less credible.'' (\cite[p. 7]{pearl_1997}, with the emphasis as in the original).

\subsubsection{A patient may suffer from two or more diseases at the same time}\label{disease_combinations}

As noted above, it is possible for a patient to suffer from two or more diseases at the same time. Given that the Old patterns in the system describe single diseases, then the system would create two or more `good' alignments, each one corresponding to one of the diseases that the patient is suffering from.

If we want the system to calculate probabilities for combinations of diseases, then the repository of Old patterns must contain patterns that represent combinations of that kind. Each such pattern may be constructed economically using references to the component diseases, in much the same way that clusters of symptoms may be referenced, as described in Section \ref{dereferencing_of_pointers}.

As with single diseases, frequency values for a combinations of diseases may be obtained from population surveys or by the judgement of medical experts. In the absence of any direct evidence of a statistical association between two or more diseases, it seems reasonable to assume that they are statistically independent. In such cases, frequency values may be derived straightforwardly via normalised values for the frequencies of occurrence of individual diseases. Whether the frequency values for combinations of diseases are measured, estimated or derived, they can be used for the calculation of $CD$ values and probabilities in exactly the same way as for single diseases.

Of course, there are so many possible combinations of diseases that it would be impossible to store information about them all. A more practical option may be to store information in Old about individual diseases and combinations of diseases that are known to have a statistical association with each other. One may assume that all other combinations of diseases are statistically independent.

\subsection{Inferences and the diagnostic cycle}

In a multiple alignment like the one shown in Fig. \ref{alignment_1}, any symbol within an Old pattern that is {\em not} matched to a symbol in New represents an inference that may be drawn from the alignment. In this example, we may infer from the alignment {\em inter alia} that the patient is likely to have a cough and a headache and that the standard treatment for influenza is required. Probabilities of these inferences can be calculated as described in \cite{wolff_1999_prob}.

If a `good' alignment makes a prediction about some marker that may be found in the patient's blood or something that may be observed in an X-ray, this may be interpreted as a suggestion to the physician that he or she should order an appropriate blood test or X-ray. If tests of that kind or other kinds of investigation are instigated as a result of the inferences drawn from preliminary alignments, the results of those investigations, together with the original symptoms, may be fed back into the system as New information. The system may then be run again and the alignments that are created may suggest a final diagnosis or the need for further investigation---and so on.

\subsection{Causal reasoning}\label{causal_reasoning}

Apart from the kinds of inference just described, medical diagnosis often seems to involve a `deeper' kind of reasoning about the causes of symptoms and diseases, using knowledge of bacteria, viruses, anatomy, physiology and so on.

The SP framework supports a variety of styles of reasoning, including probabilistic `deductive' reasoning, abductive reasoning, nonmonotonic reasoning and (as we saw in Section \ref{explaining_away}) `explaining away' (see \cite{wolff_1999_prob}). So there are reasons to believe that, within the SP framework, it may be possible to extend the pattern recognition analysis described above to incorporate causal styles of reasoning.

Recent investigation has confirmed this expectation. The input-output relations of each subsystem within a larger system can be modelled in the SP framework as a set of patterns, and causal connections can be established by matching outputs to inputs. As with the analysis described above, a `framework' pattern is also needed to ensure that alignments can be formed in an appropriate manner. These potential applications of the system need further exploration and development.

\subsection{Classes and subclasses of diseases}\label{class_hierarchy}

One of the attractions of the SP system is that it allows concepts to be represented at multiple levels of abstraction (e.g., `cat', `mammal', `vertebrate', `animal') in the manner of object-oriented design and, via the building of multiple alignments, it allows a specific entity (such as ``my cat Tibs'') to be recognised at several different levels of abstraction \cite{wolff_sp_overview,wolff_1999_prob}.

To some extent, this idea is already illustrated by the example shown in Fig. \ref{alignment_1}. The concept of `fever', represented by the pattern in column 4 of the figure, may be seen as a superclass comprising all the diseases where the patient may be feverish. Likewise, the pattern for flu symptoms (column 3 in the figure) may be seen as a superclass of the diseases in which such symptoms may be seen.

By contrast with the classification of animals and plants, the hierarchy of diseases tends to be relatively flat. However, there is scope for the recognition of classes and subclasses in the variants of diseases such as influenza and diabetes. With the SP system, each variant of a given disease may be recorded as a pattern that specifies the symptoms that are characteristic of the variant. Provided that pattern contains a symbolic link to another pattern describing the main symptoms of the disease, there is no need to repeat those symptoms redundantly in each of the variants.

\subsection{Acquisition of knowledge}\label{knowledge_acquisition}

Broadly speaking, the knowledge that is required in any artificial system for medical diagnosis can be obtained `manually' from experts or written sources, or it may be obtained by the automatic or semi-automatic abstraction of knowledge from raw medical data, or some combination of the two. The SP system has potential to facilitate any or all of these processes.

\subsubsection{Elicitation of expert knowledge}\label{knowledge_elicitation}

It should be apparent from the example described above that the SP system provides a means of representing medical knowledge in a form that is simple and intuitive. The simplicity of representing all knowledge as patterns is, perhaps, less important than the fact that this system allows computer-based knowledge to be expressed in a form that apparently reflects the natural structure of the original concepts.

This feature of the system  should facilitate traditional kinds of knowledge elicitation from experts or written sources. Medical experts should have little difficulty in expressing their knowledge directly in the form of SP patterns. Given that such experts are often busy and their time is, in any case, expensive, there are advantages if at least some of the process of building computer-based knowledge bases can be undertaken by knowledge engineers without specialised medical training. It should be possible for such people to derive a good deal of the necessary knowledge from medical text books and other written sources.

\subsubsection{Automatic or semi-automatic learning}\label{automatic_learning}

Section \ref{unsupervised_learning} presented an outline description of how the SP70 model learns the kinds of structures found in the syntax of natural languages. As was indicated in that section, that style of learning is probably not entirely appropriate for medical data but the same general principles should apply. This subsection describes in outline how the SP system may be applied to the learning of medical knowledge.

The simplest kind of `learning' is simply to keep a record of the symptoms of each specific patient and the corresponding diagnosis. This is the principle of `case-based learning' and, in conjunction with a system that kind find good partial matches between the description of a new patient and stored knowledge of old patients, it can be quite practical and useful in making new diagnoses. Since the SP system is capable of finding good partial matches between patterns, it could well be used in this way, as suggested in Section \ref{case_based_reasoning}, below.

However, medical practitioners normally recognise a degree of abstraction in medical knowledge as described in Section \ref{class_hierarchy}: a concept like `fever' describes a cluster of symptoms that is found in two or more different diseases; diseases like influenza often come in several variants or subclasses; and symptoms may be described in terms of ranges rather than specific values as in our example of temperature (Section \ref{uncertainties_in_diagnosis}).

As an illustration of the way in which the SP system may create these kinds of abstraction, consider two imaginary patients with two different diseases and symptoms represented by the patterns `X A B Y C D E Z F' and `A P Q B C R D S E F T'. With these two patterns, the SP system would create an alignment like this:

\begin{center}
\begin{BVerbatim}
0 X A     B Y C   D   E Z F   0
    |     |   |   |   |   |
1   A P Q B   C R D S E   F T 1,
\end{BVerbatim}
\end{center}

\noindent and from the matched symbols in this alignment it may derive the pattern `A B C D E F'. With the addition of some appropriate ID-symbols, this pattern may serve like the pattern for `fever' in Figs. \ref{patterns_1} and \ref{alignment_1}: it represents a cluster of symptoms that appears in two or more different diseases. Alternatively, this pattern may represent the symptoms of a general class of diseases with two variants represented by the two original patterns, modified so that the shared cluster of symptoms are replaced by a pointer to the general class.

With regard to categories like the set of two alternative temperatures represented by the patterns `$<$t1$>$ 38-39 $<$/t1$>$' and `$<$t1$>$ 40+ $<$/t1$>$' in Fig. \ref{patterns_1}, the SP system may derive this kind of disjunctive category in much the same manner as the disjunctive class \{`b o y', `g i r l'\} in the example in Section \ref{unsupervised_learning}.

\subsection{Integration}

The very simple format for representing knowledge described in Section \ref{representation_of_knowledge} is intended to be as nearly `universal' as possible in the sense that it is designed to represent a wide range of different kinds of knowledge. This should be much more nearly true when the concept of {\em pattern} has been generalised to two dimensions.

In a similar way, the concept of multiple alignment described in Section \ref{processing_knowledge} is intended to be a `universal' model for a wide range of different kinds of processing: pattern recognition, information retrieval, various kinds of reasoning, and so on.

To the extent that these two objectives can be realised, they should facilitate the seamless integration of different kinds of application, including medical diagnosis.  It should, for example, be relatively easy to apply a natural language interface to an SP system for medical diagnosis or to integrate the kind of visual pattern recognition needed in the diagnosis of different kinds of skin cancer with other kinds of medical expertise.

\section{Comparison with alternatives}\label{alternatives_section}

As mentioned in the introduction, a wide variety of philosophies and systems have been applied to the problem of medical diagnosis. In this section, I briefly review some of the more prominent of these approaches and compare them with the SP approach, as described in this paper.

\subsection{Rule-based systems}

Rule-based systems (like the well-known MYCIN system \cite{shortliffe_1976}) contain if-then rules where the `if' side of any rule is a collection of one or more conditions for the rule to fire connected by logical operators such as `AND', `OR' (which may be inclusive or exclusive) and `NOT'. By contrast, the SP system expresses all knowledge in the form of patterns.

At first sight, SP patterns lack the expressive power of if-then rules. But the effect of such rules can be modelled within the SP system if that is required \cite{wolff_1999_comp,wolff_maths_logic}. And if medical diagnosis is viewed as a process of pattern recognition (as in this paper), then SP patterns and the SP framework are, arguably, a more natural and flexible medium for the representation and processing of knowledge than are if-then rules.

To illustrate this last point, we can express the distinctive features of influenza by a rule such as:

\begin{center}
\fontsize{08.00pt}{09.60pt}
\begin{BVerbatim}
IF chills AND cough AND headache AND aching muscles AND runny nose AND sore throat
     THEN influenza (probability = 0.9)
\end{BVerbatim}
\end{center}

Although there may be a probability associated with the rule (as shown), the rule has an intrinsic logic which, if strictly applied, means that the rule will {\em only} fire if all the conditions are satisfied. By contrast, a pattern like the one shown in column 3 of Fig. \ref{alignment_1} may appear in the best alignment when any reasonably large subset of its symbols have been matched.

If one attempted to achieve this kind of flexibility with an if-then rule using combinations of AND, OR and NOT, the rule would become very complex. Alternatively, one might split up the rule into a number of smaller rules, one for each symptom or combination of two or three symptoms---but again the result would be relatively complex.

\subsubsection{Probabilities}

In systems like MYCIN and some of its successors, the `probabilities' that the system calculates are really measures of confidence without the theoretical underpinnings of probability theory. In other systems, ``... formal approaches based on probability theory are precise but can be awkward and non-intuitive to use.'' \cite[p. 272]{fox_etal_2001}. By contrast, the SP framework allows true probabilities to be calculated quite simply (see Section \ref{calculation_of_probabilities} and \cite{wolff_1999_prob}) and strictly in accordance with established theory (as described in sources such as \cite{cover_thomas_1991}).

\subsection{Neural networks}

One of the attractions of artificial neural networks for the support of medical diagnosis is that they can be trained with appropriate data, thus by-passing the need for the manual compilation of knowledge by medical experts or knowledge engineers. However, ``A major drawback is that `knowledge' embedded [in the neural network] is cryptically coded as a large number of weights and activation values. As a consequence, the lack of neural network validation tools is often one of the reasons limiting their use in practice, especially in the context of medical diagnosis where physicians cannot trust a system without explanation of its decisions.'' \cite[pp. 141--142]{bologna_2003}.

While there may be scope for extracting rules from a trained neural network ({\em ibid.}), this adds complexity and uncertainty to the technology and defeats the other main attraction of a neural network: as a classifier of specific cases in terms of the learned knowledge.

As a system for unsupervised learning of knowledge structures from raw data, the SP system is not yet a rival to existing neural network systems. However, the system has clear potential for unsupervised learning and, if that potential can be realised, the system has the advantage that its knowledge is stored in a form that can be read and understood by people. Meanwhile, if it is supplied with knowledge about diseases derived from experts or text books, it can be used for diagnostic classification of individual patients.

\subsection{Fuzzy logic and fuzzy set theory}

Given the variability of diseases and other uncertainties associated with medical diagnosis (Section \ref{uncertainties_in_diagnosis}), the field of fuzzy logic (and fuzzy set theory) has the obvious attraction that it has been designed with the explicit intention of providing a model for `fuzzy' concepts and `fuzzy' operations on them (see, for example, \cite{klir_yuan_1995,chen_1994}).

In purely theoretical terms, the field of fuzzy logic may be criticised because it introduces a fairly elaborate conceptual framework to accommodate the undoubtedly fuzzy nature of many human concepts but this conceptual framework is poorly integrated with other ideas about the nature of human cognition. By contrast, the SP theory grew out of research in psychology and it provides a unified model for several aspects of human perception and cognition \cite{wolff_sp_overview}.

Considerations of that kind may be discounted as not relevant to the practicalities of medical diagnosis. But in that connection fuzzy logic has the drawback that it introduces another layer of complexity to the already difficult process of eliciting knowledge from medical experts \cite{boegl_etal_2004}. There seems to be some scope for ameliorating this problem by the provision of appropriate tools ({\em ibid.}) but the basic problem remains. By contrast, the SP system allows concepts to be expressed in a simple, intuitive manner and, at the same time, it accommodates much, perhaps all, of the fuzziness of medical diagnosis.

\subsection{Bayesian networks}

Two of the main differences between Bayesian networks (see, for example, \cite{pearl_1997}) and the SP system are:

\begin{itemize}

\item Bayesian networks focus on the binary relationship between any given node in the network and each of its parent nodes (if any). In this respect, they inherit some of the thinking behind if-then rules. By contrast, the SP system is oriented towards the representation and processing of associations (expressed as patterns) that may contain arbitrarily many elements.

\item Correspondingly, any given Bayesian network stores its statistical knowledge in the form of tables of conditional probabilities, one for each node in the network. By contrast, the SP system stores its statistical knowledge in the form of integers, one for each pattern, representing the absolute or relative frequency of that pattern in some domain.

\end{itemize}

These and related differences seem to underlie some of the apparent drawbacks of Bayesian networks compared with the SP framework:

\begin{itemize}

\item The directional nature of Bayesian networks does not sit easily with the non-directional nature of medical syndromes.

\item The process of calculating probabilities of inferences in a Bayesian network is substantially more complicated than the calculation of probabilities for alignments and inferences in the SP framework.

\item The tables of conditional probabilities required for Bayesian networks are significantly more complex than simple measures of frequency that are used in the SP system. Notwithstanding the development of special methods for eliciting conditional probabilities from experts \cite{van_der_gaag_etal_2002}, the process of building up the necessary tables of conditional probabilities is likely to be much harder than measuring or estimating an integer value for each disease, reflecting its absolute or relative frequency in a given domain.

\end{itemize}

\subsection{Case-based reasoning}\label{case_based_reasoning}

A major attraction of case-based reasoning in medical diagnosis (see, for example, \cite{schmidt_gierl_2001,althoff_etal_1998}) is that, compared with many of the alternatives, it can considerably simplify the process of acquiring the necessary knowledge. In its simplest form, a case-based system merely requires a description of one or more specific examples of each disease and a search algorithm that can find exact matches or good partial matches between the symptoms of a given patient and one or more of the stored records.

In some respects, the SP system is like a case-based system and it could indeed be used like a case-based system. To use it in this way, each of the Old patterns should represent a specific case (including its diagnosis) and the New pattern or patterns should represent the symptoms of a patient for whom a diagnosis is required. The capabilities of the system for finding exact matches and good partial matches between patterns will allow it to retrieve patterns for previously-diagnosed cases that are similar to any given current case.

The main advantages of the SP system compared with the case-based approach to diagnosis are:

\begin{itemize}

\item It facilitates the description of diseases in generalised terms without the need to specify exact values for every category of symptom. In our main example, we saw how the temperature of a patient with a disease like influenza may be specified as a range of alternative values (Section \ref{uncertainties_in_diagnosis}). Any other category of symptom may be treated in the same way.

\item It allows one to specify clusters of symptoms that are found in two or more different diseases and it allows one to describe diseases at two or more levels of abstraction (Section \ref{class_hierarchy}). Both of these things facilitate the description of diseases without the need to repeat information unnecessarily where similar patterns are found in different diseases or varieties of disease.

\end{itemize}

\section{Conclusion}\label{conclusion_section}

As we have seen, the SP system accommodates the main elements of medical diagnosis, viewed as a problem of pattern recognition, and there are reasons to believe that it may also provide support for causal reasoning in medical diagnosis. However, the SP62 model is only a prototype that serves for research and demonstration. It is not yet a system with `industrial strength'. The main developments that are needed to reach that goal are:

\begin{itemize}

\item The provision of a well-designed graphical user interface.

\item There is probably scope for improvements in the search methods that are used within the system.

\item There is scope for the application of parallel processing both to improve the scaling properties of the system (Section \ref{computational_complexity}) and to increase absolute speeds of processing.

\item Naturally, the system needs to be provided with appropriate knowledge. For each area of application, a set of patterns needs to be developed that describes the diseases and symptom clusters in that domain.

\item At some stage after the development of a realistic knowledge base, the performance of the system must be validated against the judgement of human medical experts.

\end{itemize}

The potential payoff from these developments is a system that allows knowledge about diseases to be expressed in a simple, intuitive manner, that can cope with errors and uncertainties in knowledge about diseases and knowledge about individual patients, that simplifies the acquisition and storage of statistical information, that calculates true probabilities of diagnoses, that smooths the path to the automatic or semi-automatic abstraction of medical knowledge in the future, and should facilitate the integration of medical diagnosis with other kinds of application.

\clearpage

\begin{figure}[!hbt]
\includegraphics[width=0.25\textwidth]{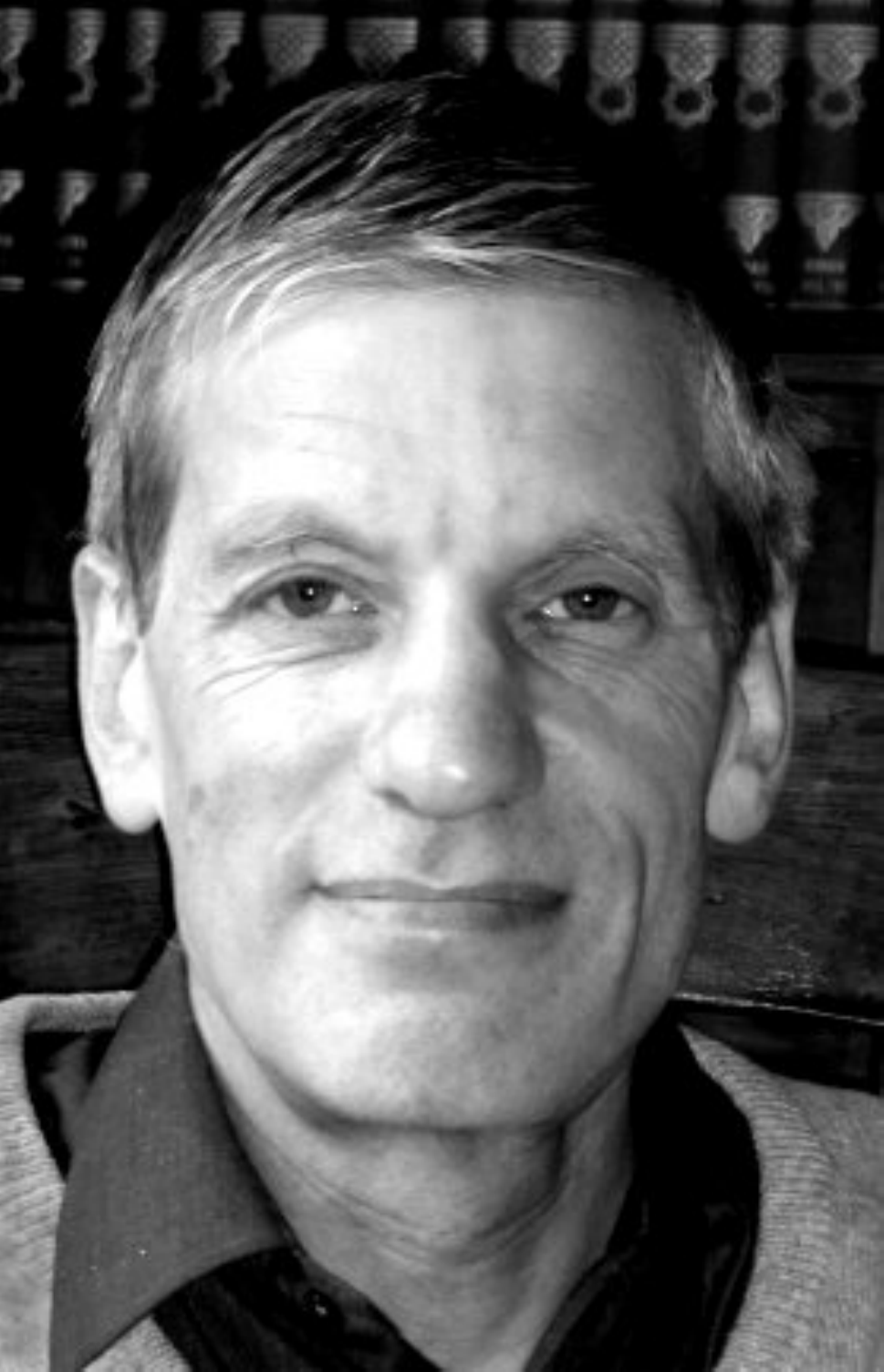}
\end{figure}

Gerry Wolff is Director of CognitionResearch.org.uk. Previously, he has held academic posts in the School of Informatics, University of Wales, Bangor, and the Department of Psychology, University of Dundee, a Research Fellowship with IBM in Winchester, UK, and he has worked for four years as a Software Engineer with Praxis Systems in Bath, UK. His first degree is in Natural Sciences (Psychology) from Cambridge University and his PhD is in Cognitive Science from the University of Wales. He is a Chartered Engineer and Member of the British Computer Society. Since 1987 his research has focussed on the development of the SP theory. Previously his main research interests were in developing computer models of language learning. He has numerous publications in a wide range of journals, collected papers and conference proceedings.

\end{document}